\documentclass{article}
\usepackage{ijcai18}

\usepackage{times}
\usepackage{xcolor}
\usepackage{soul}
\usepackage[utf8]{inputenc}
\usepackage[small]{caption}

\usepackage{amsmath}
\usepackage{graphicx}
\usepackage[linesnumbered,ruled]{algorithm2e}
\usepackage{hyperref}

\makeatletter
\DeclareRobustCommand\onedot{\futurelet\@let@token\@onedot}
\def\@onedot{\ifx\@let@token.\else.\null\fi\xspace}

\def\ie{\emph{i.e}\onedot}

\makeatother

\title{MSC: A Dataset for Macro-Management in StarCraft II}

\author{
Huikai Wu, 
Yanqi Zong\thanks{Independent Researcher, zongyanqi@gmail.com},
Junge Zhang, 
Kaiqi Huang
\\ 
NLPR, Institute of Automation, Chinese Academy of Sciences \\
\{huikai.wu, jgzhang, kaiqi.huang\}@nlpr.ia.ac.cn \\
\\
\textbf{Homepage: \hyperref[]{https://github.com/wuhuikai/MSC}}
}

\begin{document}

\maketitle

\begin{abstract}
  Macro-management is an important problem in StarCraft, which has been studied for a long time.
Various datasets together with assorted methods have been proposed in the last few years.
But these datasets have some defects for boosting the academic and industrial research:
1) There are neither standard preprocessing, parsing and feature extraction procedure nor predefined training, validation and test set in some datasets.
2) Some datasets are only specified for certain tasks in macro-management.
3) Some datasets either are too small or don't have enough labeled data for modern machine learning algorithms such as deep learning.
Thus, most previous methods are trained with various features, evaluated on different test sets from the same or different datasets, making it difficult to be compared directly.
To boost the research of macro-management in StarCraft, we release a new dataset MSC based on the platform SC2LE.
MSC consists of well-designed feature vectors, predefined high-level actions and final result of each match.
We also split MSC into training, validation and test set for the convenience of evaluation and comparison.
Besides the dataset, we propose baseline models and present initial baseline results for global state evaluation and build order prediction, which are two of the key tasks in macro-management.
\end{abstract}

\section{Introduction}
\begin{figure*}
\begin{center}
	\includegraphics[width=\linewidth]{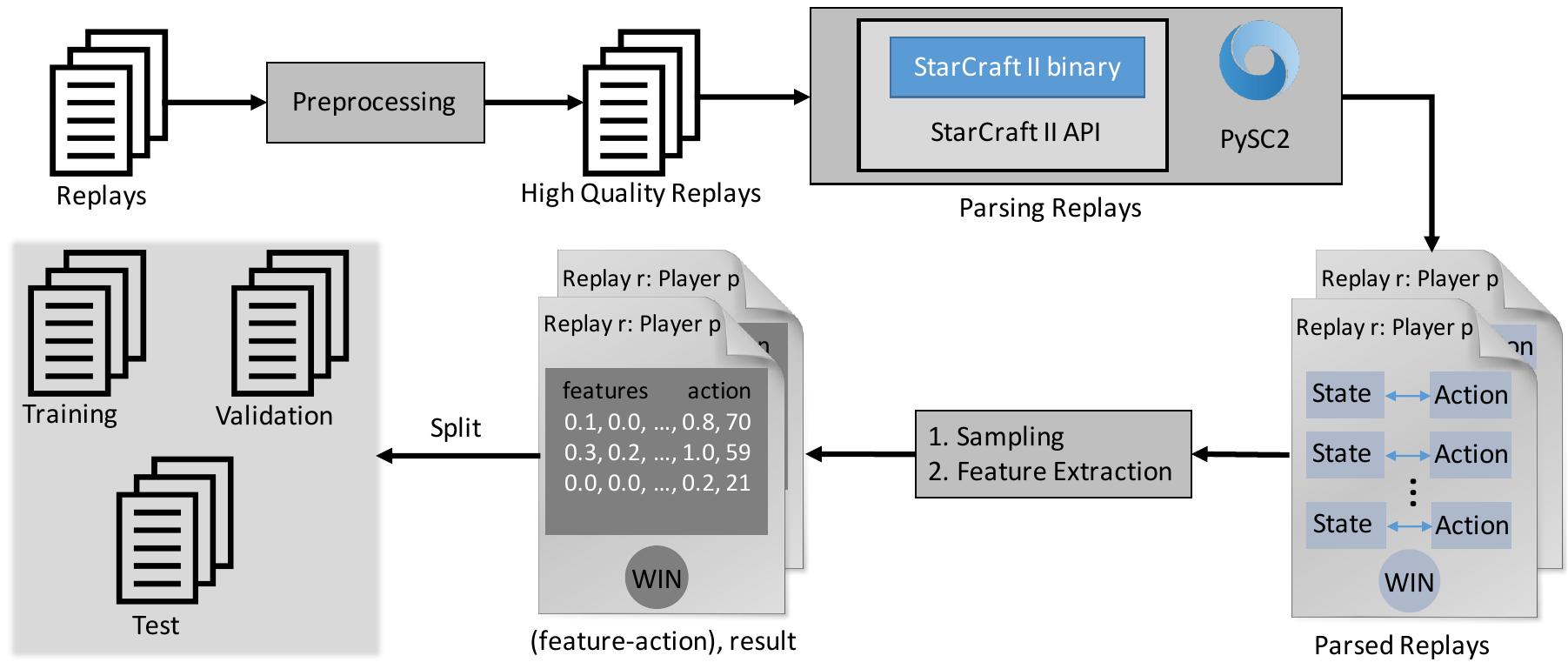}
\end{center}
	\caption{\textbf{Framework overview of MSC.}
	Replays are firstly filtered according to predefined criteria and then parsed with PySC2.
	The states in parsed replays are sampled and turned into feature vectors.
	The final files which contain feature-action pairs and the final results are split into training, validation and test set.}
	\label{fig:msc}
\end{figure*}
Deep learning has surpassed the previous state-of-the-art in playing Atari games~\cite{mnih2015human}, the classic board game Go~\cite{silver2016mastering} and the 3D first-person shooter game Doom~\cite{lample2017playing}.
But it remains as a challenge to play real-time strategy (RTS) games like StarCraft II with deep learning algorithms~\cite{vinyals2017starcraft}.
Such games usually have enormous state and action space compared to Atari games and Doom.
Furthermore, RTS games are usually partially observable, in contrast to Go.

Recent experiment has shown that it's difficult to train a deep neural network (DNN) end-to-end for playing StarCraft II.
\cite{vinyals2017starcraft} introduce a new platform SC2LE on StarCraft II and train a DNN with Asynchronous Advantage Actor Critic (A3C)~\cite{mnih2016asynchronous}.
Unsurprisingly, the agent trained with A3C couldn't win a single game even against the easiest built-in AI.
Based on this experiment and the progresses made in StarCraft I such as micro-management~\cite{peng2017multiagent}, build order prediction~\cite{justesen2017learning} and global state evaluation~\cite{erickson2014global}, we believe that treating StarCraft II as a hierarchical learning problem and breaking it down into micro-management and macro-management is a feasible way to boost the performance of current AI bots.

Micro-management includes all low-level tasks related to unit control, such as collecting mineral shards and fighting against enemy units;
while macro-management refers to the higher-level game strategy that the player is following, such as build order prediction and global state evaluation. 
We could obtain near-human performance in micro-management easily with deep reinforcement learning algorithms such as A3C~\cite{vinyals2017starcraft},
while it's hard to solve macro-management at present, though lots of efforts have been made by StarCraft community~\cite{churchill2011build,synnaeve2011bayesian,erickson2014global,justesen2017learning}.
One promising way for macro-management is to mimic professional human players with machine learning methods.
\cite{erickson2014global} learn to evaluate the global state from replays while~\cite{justesen2017learning} utilize DNN for build order prediction.
Both methods are trained with replays, which are official log files used to record the entire game status when playing StarCraft.

There are many datasets released in StarCraft I for learning macro-management from replays~\cite{weber2009data,cho2013replay,erickson2014global,justesen2017learning}.
But these datasets are designed for specific tasks in macro-management and didn't release the partition for training, validation and test set.
Besides, datasets in~\cite{cho2013replay,erickson2014global} only contain about 500 replays, which are too small for modern machine learning algorithms.
StarData~\cite{lin2017stardata} is the largest dataset in StarCraft I, containing 65646 replays.
But there are only a few replays labeled with the final results, which is not suitable for many tasks in macro-management, such as global state evaluation.
SC2LE~\cite{vinyals2017starcraft} is the largest dataset in StarCraft II, which has 800K replays.
However, there is neither a standard processing procedure nor predefined training, validation and test set.
Besides, it's designed for end-to-end human-like control of StarCraft II, which is not easy to use for tasks in macro-management.

Besides a standard dataset, macro-management algorithms could also be compared by building AI bots which differ only in the macro-management algorithm to play against each other. 
However, such a method works as a black box in its nature. The reasons why an algorithm wins are mixed and uncertain.
For example, the winning algorithm might be trained with a much larger dataset; the dataset might be composed of replays collected from more professional players; the features might contain more information; or, the macro-management module might be more compatible with the specific AI bot.
It is hard to tell which macro-management algorithm is better unless a standard dataset is used.

To take the research of learning macro-management from replays a step further, we build a new dataset MSC based on SC2LE.
It's the biggest dataset dedicated for macro-management in StarCraft II, which could be used for assorted tasks like build order prediction and global state evaluation.
We define standard procedure for processing replays from SC2LE, as shown in Figure~\ref{fig:msc}.
After processing, our dataset consists of well-designed feature vectors, pre-defined action space and the final result of each match.
All replays are then divided into training, validation and test set.
Based on MSC, we train baseline models and present the initial baseline results for global state evaluation and build order prediction, which are two of the key tasks in macro-management.
Our main contributions are two folds and summarized as follows:
\begin{itemize}
  \item We build a new dataset MSC for macro-management on StarCraft II, which contains standard preprocessing, parsing and feature extraction procedure. The dataset is divided into training, validation and test set for the convenience of evaluation and comparison between different methods.
  \item We propose baseline models together with initial baseline results for two of the key tasks in macro-management i.e. global state evaluation and build order prediction.
\end{itemize}
\section{Related Work}
We briefly review the related works of macro-management in StarCraft.
Then, we compare our dataset with several released datasets which are suitable for macro-management.
\subsection{Macro-Management in StarCraft}
We introduce some background for StarCraft I and StarCraft II shortly, and then review several related works focusing on various tasks in macro-management.
\paragraph{StarCraft}
StarCraft I is a RTS game released by Blizzard in 1998.
In the game, each player controls one of three races including Terran, Protoss and Zerg to simulate a strategic military combat.
The goal is to gather resources, build buildings, train units, research techniques and finally, destroy all enemy buildings.
During playing, the areas which are unoccupied by friendly units and buildings are unobservable due to the fog-of-war, which makes the game more challenging.
The players must not only control each unit accurately and efficiently but also make some strategic plans given current situation and assumptions about enemies.
StarCraft II is the next generation of StarCraft I which is better designed and played by most StarCraft players.
Both in StarCraft I and StarCraft II, build refers to the union of units, buildings and techniques.
Order and action are interchangeably used which mean the controls for the game.
Replays are used to record the sequence of game states and actions during a match, which could be watched from the view of enemies, friendlies or both afterwards.
There are usually two or more players in a match, but we focus on the matches that only have two players, noted as enemy and friendly.
\paragraph{Macro-Management}
In StarCraft community, all tasks related to unit control are called micro-management, while macro-management refers to the high-level game strategy that the player is following.
Global state evaluation is one of the key tasks in macro-management, which focuses on predicting the probability of winning given current state~\cite{erickson2014global,stanescu2016evaluating,ravari2016starcraft,sanchez2017machine}.
Build order prediction is used to predict what to train, build or research in next step given current state~\cite{hsieh2008building,churchill2011build,synnaeve2011bayesian,justesen2017learning}.
\cite{churchill2011build} applied tree search for build order planning with a goal-based approach.
\cite{synnaeve2011bayesian} learned a Bayesian model from replays while~\cite{justesen2017learning} exploited DNN.
Opening strategy prediction is a subset of build order prediction, which aims at predicting the build order in the initial stage of a match~\cite{kostler2013multi,blackford2014real,justesen2017continual}.
\cite{dereszynski2011learning} work on predicting the state of enemy while~\cite{cho2013replay} try to predict enemy build order.
\subsection{Datasets for Macro-Management in StarCraft}
There are various datasets for macro-management, which could be subdivided into two groups.
The datasets in the first group usually focus on specific tasks in macro-management, while the datasets from the second group could be generally applied to assorted tasks.
\paragraph{Task-Oriented Datasets}
The dataset in~\cite{weber2009data} is designed for opening strategy prediction. 
There are 5493 replays of matches between all races, while our dataset contains 36619 replays.
\cite{cho2013replay} learns to predict build order with a small dataset including 570 replays in total.
\cite{erickson2014global} designed a procedure for preprocessing and feature extraction among 400 replays.
However, these two datasets are both too small and not released yet.
\cite{justesen2017learning} also focuses on build order prediction and builds a dataset containing 7649 replays.
But there are not predefined training, validation and test set.
Compared to these datasets, our dataset is more general and much larger, besides the standard processing procedure and dataset division.
\paragraph{General-Purpose Datasets}
The dataset proposed in~\cite{synnaeve2012dataset} is widely used in various tasks of macro-management.
There are 7649 replays in total but barely with the final result of a match.
Besides, it also lacks a standard feature definition, compared to our dataset.
StarData~\cite{lin2017stardata} is the biggest dataset in StarCraft I containing  65646 replays.
However, it's not suitable for tasks that require the final result of a match, as most replays don't have the result label.
\cite{vinyals2017starcraft} proposed a new and large dataset in StarCraft II containing 800K replays.
We transform it into our dataset for macro-management with standard processing procedure, well-designed feature vectors, pre-defined high-level action space as well as the division of training, validation and test set.
\section{Dataset}
\label{section:dataset}
Macro-management in StarCraft has been researched for a long time, but there isn't a standard dataset available for evaluating various algorithms.
Current research on macro-management usually needs to collect replays firstly, and then parse and extract hand-designed features from the replays, which causes that there is neither unified datasets nor consistent features.
As a result, nearly all the algorithms in macro-management couldn't be compared with each other directly.

We try to build a standard dataset MSC, 
which is dedicated for macro-management in StarCraft II, with the hope that it could serve as the benchmark for evaluating assorted algorithms in macro-management.
To build our dataset, we design a standard procedure for processing the 64396 replays, as shown in Figure~\ref{fig:msc}.
We first preprocess the replays to ensure their quality.
We then parse the replays using PySC2\footnote{https://github.com/deepmind/pysc2}.
We sample and extract feature vectors from the parsed replays subsequently and then divide all the replays into training, validation and test set.
\begin{table}
\begin{center}
\begin{tabular}{|c|c|c|c|c|c|c|}
\hline
V.S. & TvT & TvP & TvZ & PvP & PvZ & ZvZ \\
\hline
\#Replays & 4897 & 7894 & 9996 & 4334 & 6509 & 2989 \\
\hline
\end{tabular}
\end{center}
\caption{\textbf{The number of replays after applying our pipeline.}}
\label{table:replays}
\end{table}
\subsection{Preprocessing}
There are 64396 replays in SC2LE, which could be split into 6 groups according to the races in the matches.
To ensure the quality of the replays in our dataset, we drop out all the replays dissatisfying the criteria:
\begin{itemize}
  \item Total frames of a match must be greater than 10000.
  \item The APM (Actions Per Minute) of both players must be higher than 10.
  \item The MMR (Match Making Ratio) of both players must be higher than 1000.
\end{itemize}
Because low APM means that the player is standing around while low MMR refers to a broken replay or a player who is weak.

After applying these criteria, we obtain 36619 high quality replays, which are unbroken and played by relatively professional players.
The number of replays in each group after preprocessing is summarized in Table~\ref{table:replays}.
\subsection{Parsing Replays}
\label{subsection:parse}
\paragraph{Build Order Space}
We define a high-level action space $A$, which consists of four groups: Build a building, Train a unit, Research a technique and Morph (Update) a building\footnote{Cancel, Halt and Stop certain actions from $A$ are also included for completion.}.
We also define an extra action $a_\emptyset$, which means doing nothing.
Both $A$ and $a_\emptyset$ constitute the entire build order space.
\paragraph{Observation Definition}
Each observation we extract includes (1) buildings, units and techniques owned by the player, (2) resources used and owned by the player and (3) enemy units and buildings which are observed by the player.
\paragraph{Parsing Process}
The preprocessed replays are parsed using Algorithm~\ref{algo:parse_replay} with PySC2, which is a python API designed for reading replays in StarCraft II.
When parsing replays, we extract an observation $o_t$ of current state and an action set $A_t$ every $n$ frames, where $A_t$ contains all actions since $o_{t-1}$.
The first action in $A_t$ that belongs to $A$ is set to be the target build order for observation $o_{t-1}$.
If there's no action belonging to $A$, we take $a_\emptyset$ as the target.
When reaching the end of a replay, we save all (observation, action) pairs and the final result of the match into the corresponding local file.
$n$ is set to be $8$ in our experiments, because in most cases, there's at most one action belonging to $A$ every 8 frames.
\begin{algorithm}[t]
\caption{Replay Parser}
\label{algo:parse_replay}
Global: \textbf{List} states = []

Global: \textbf{Observation} previousObservation = None

\While{True}{
\textbf{Observation} currentObservation $\leftarrow$ observation of current frame

\textbf{List} actions $\leftarrow$ actions conducted since previousObservation

\textbf{Action} action = $a_\emptyset$

\For{a in actions}{
\If{$a \in \{Build, Train, Research, Morph\}$}{
action = a

break
}
}

states.append((previousObservation, action))

previousObservation $\leftarrow$ currentObservation

\If{reach the end of the replay}{
\textbf{Result} result $\leftarrow$ result of this match (win or lose)

\Return (result, states)
}
Skip $n$ frames
}
\end{algorithm}
\begin{figure}[t]
\begin{center}
	\includegraphics[width=\linewidth]{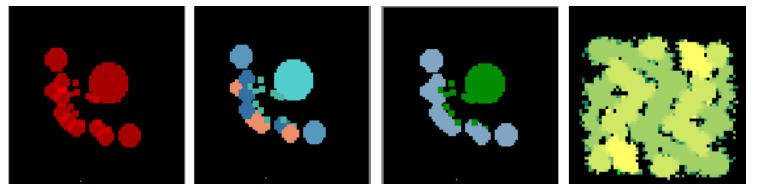}
\end{center}
	\caption{\textbf{Visualization of spatial feature tensor.} From left to right, the images represent unit density, unit type, unit relative and map height respectively.}
	\label{fig:spatial}
\end{figure}
\subsection{Sampling and Extracting Features} 
\label{subsection:sampling}
The number of action $a_\emptyset$ is much larger than the total number of high-level actions in $A$.
Thus, we sample the (observation, action) pairs in the parsed files to balance the number of these two kinds of actions, and then extract features from them.
The features we extracted are composed of global feature vectors and spatial feature tensors.
The global feature vector is normalized into [0, 1], which consists of a few sub-vectors described here in order:
\begin{enumerate}
  \item frame id.
  \item the resources collected and used by the player.
  \item the alerts received by the player.
  \item the upgrades applied by the player.
  \item the techniques researched by the player.
  \item the units and buildings owned by the player.
  \item the enemy units and buildings observed by the player.
\end{enumerate}
The spatial feature tensor is roughly normalized into [0, 1] with shape $13\times 64\times 64$, which contains similar features to global feature vector with additional spatial information, as shown in Figure~\ref{fig:spatial}. 

Once features are extracted, we split our dataset into training, validation and test set in the ratio 7:1:2.
The ratio between winners and losers preserves 1:1 in the three sets.
\subsection{Downstream Tasks}
\begin{figure}[t]
 \begin{center}
 	\includegraphics[width=\linewidth]{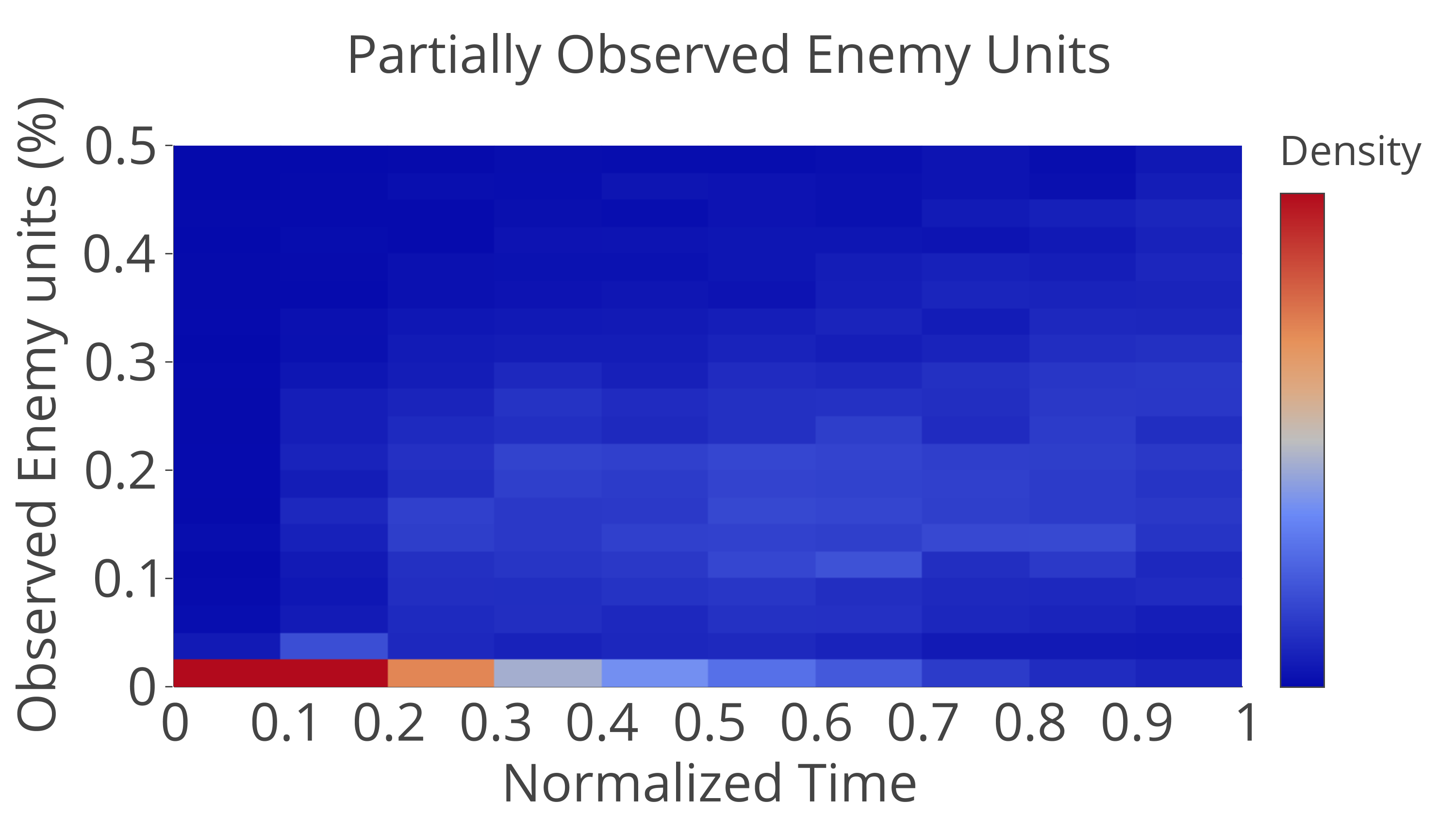}
 \end{center}
 	\caption{\textbf{Density of partially observed enemy units.} X-axis represents the progress of the game while Y-axis is the ratio between the number of partially observed enemy units and total enemy units. Best viewed in color.}
 	\label{fig:po}
 \end{figure}

Our dataset MSC is designed for macro-management in StarCraft II.
We will list some tasks of macro-management that could benefit from our dataset in this section.
 \paragraph{Long Sequence Prediction}
 Each replay is a time sequence containing states (feature vectors), actions and the final result.
 One possible task for MSC is sequence modeling.
 The replays in MSC usually have 100-300 states, which could be used for evaluting sequence models like LSTM~\cite{hochreiter1997long} and NTM~\cite{graves2014neural}.
 As for tasks in StarCraft II, MSC could be used for build order prediction~\cite{justesen2017learning}, global state evaluation~\cite{erickson2014global} and forward model learning.
 \paragraph{Uncertainty Modeling}
 Due to ``the fog of war", the player in StarCraft II could only observe friendly builds and part of enemy builds, which increases the uncertainty of making decisions.
 As shown in Figure~\ref{fig:po}, it's hard to observe enemy builds at the beginning of the game.
 Though the ratio of observed enemy builds increases as game progressing, we still know nothing about more than half of the enemy builds.
 This makes our dataset suitable for evaluating generative models such as variational autoencoders~\cite{kingma2013auto}.
 Some macro-management tasks in StarCraft such as enemy future build prediction~\cite{dereszynski2011learning} or enemy state prediction can also benefit from MSC.



\paragraph{Reinforcement Learning}
Sequences in our dataset MSC are usually more than 100 steps long with only the final 0-1 result as the reward.
It's useful to learn a reward function for every state through inverse reinforcement learning (IRL)~\cite{abbeel2004apprenticeship}, so that the AI bots can control the game more accurately.
\paragraph{Planning and Tree Search}
Games with long time steps and sparse rewards usually benefit a lot from planning and tree search algorithms.
One of the most successful applications is AlphaGO~\cite{silver2016mastering}, which uses 
Monte Carlo tree search~\cite{coulom2006efficient,kocsis2006bandit} to boost its performance.
MSC is a high-level abstraction of StarCraft II, which could be viewed as a planning problem.
Once a good forward model and an accurate global state evaluator are learned, MSC is the right dataset for testing various planning algorithms and tree search methods.
\section{Baselines for Global State Evaluation and Build Order Prediction}
\label{sec:gse}
MSC is a general-purpose dataset for macro-management in StarCraft II, which could be used for various high-level tasks.
We present the baseline models and initial baseline results for global state evaluation and build order prediction in this paper, and leave baselines of other tasks as our future work.
\subsection{Global State Evaluation}
\begin{figure}[t]
\begin{center}
	\includegraphics[width=\linewidth]{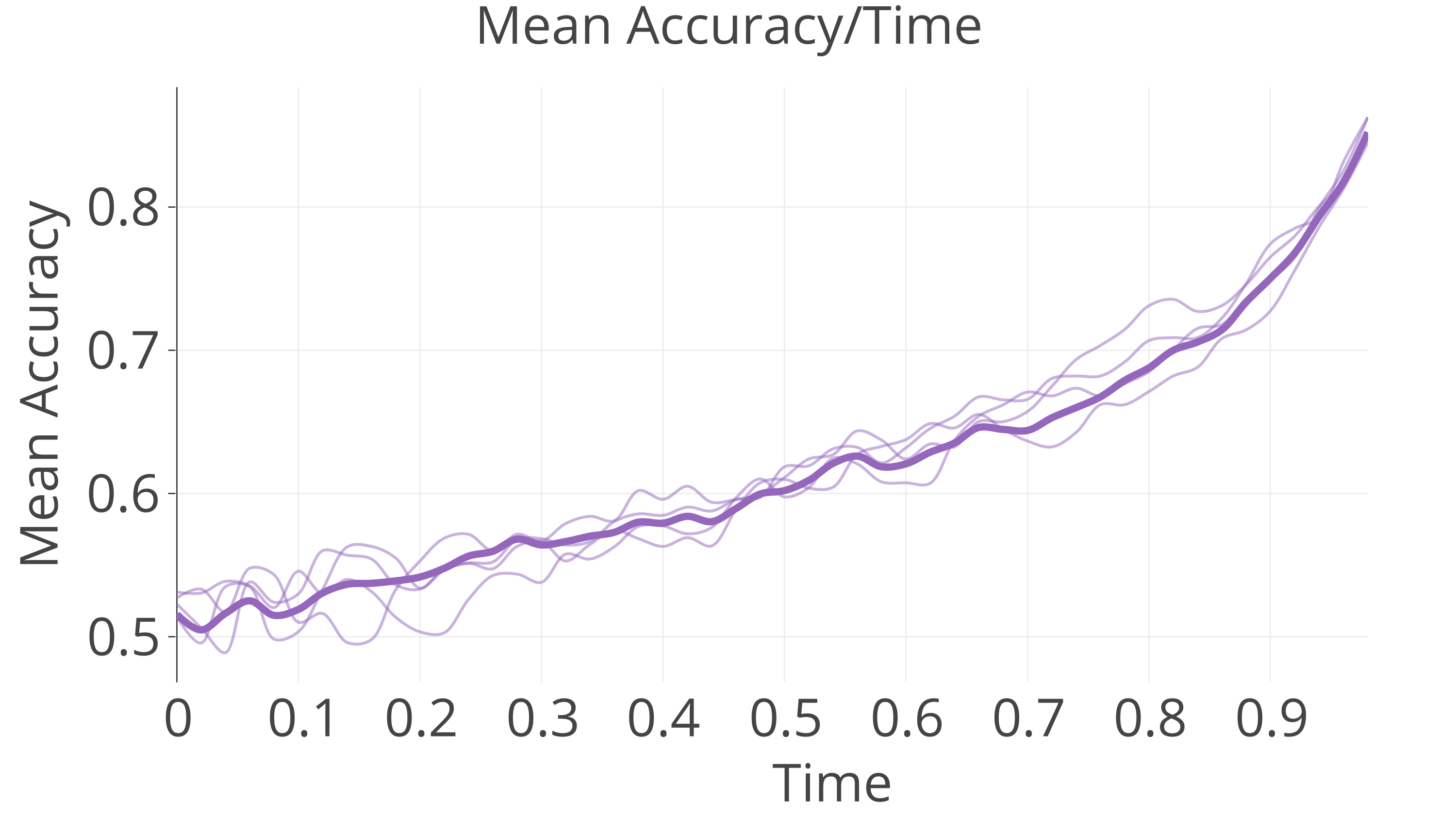}
\end{center}
	\caption{\textbf{The trend of mean accuracy with time steps for global state evaluation.} The mean accuracy on test set increases as game progresses.}
	\label{fig:map}
\end{figure}

\paragraph{Definition}
When human players play StarCraft II, they usually have a sense of whether they would win or lose in the current state.
Such a sense is essential for the decision making of what to train or build in the following steps.
For AI bots, it's also desirable to have the ability of predicting the probability of winning in a certain state.
Such an ability is called global state evaluation in StarCraft community.
Formally, global state evaluation is predicting the probability of winning given current state at time step $t$, \ie predicting the value of $P(R=win|x_t)$.
$x_t$ is the state at time step $t$ while $R$ is the final result.
Usually, $x_t$ couldn't be accessed directly, what we obtain is the observation of $x_t$ noted as $o_t$.
Thus, we use $o_1, o_2, ..., o_t$ to represent $x_t$ and try to learn a model for predicting $P(R=win|o_1, o_2, ..., o_t)$ instead.
\begin{figure}[t]
\begin{center}
	\includegraphics[width=\linewidth]{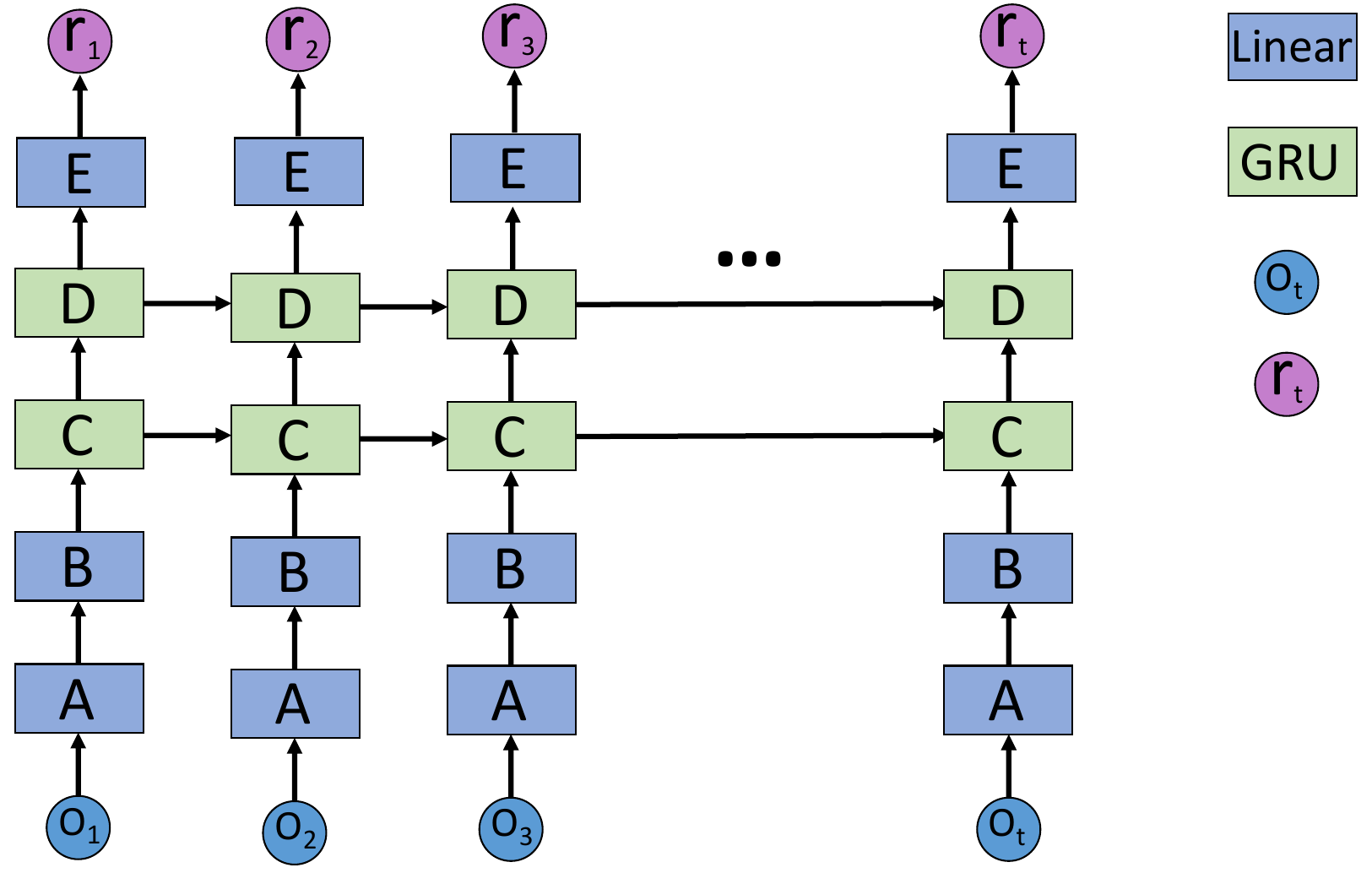}
\end{center}
	\caption{\textbf{Baseline network architecture.} $o_t$ is the input feature vector. A, B and E are linear units with the number of units 1024, 2048 and 1, while C and D are GRUs with size 2048 and 512.}
	\label{fig:baseline}
\end{figure}
\paragraph{Baseline Network}
We model global state evaluation as a sequence decision making problem and use Recurrent Neural Networks (RNNs)~\cite{mikolov2010recurrent} to learn from replays.
Concretely, we use GRU~\cite{cho2014properties} in the last two layers to model the time series $o_1, o_2, ..., o_t$.
As shown in Figure~\ref{fig:baseline}, the feature vector $o_t$ flows through linear units A and B with size 1024 and 2048.
Then two GRUs C and D with size 2048 and 512 are applied.
The hidden state from D is fed into the linear unit E followed by a Sigmoid function to get the final result $r_t$.
ReLUs are applied after both A and B.
\paragraph{Objective Function}
Binary Cross Entropy Loss (BCE) serves as our objective function, which is defined as Equation~\ref{equation:bce},
		\begin{equation}\label{equation:bce}
		\begin{aligned}
			J(\Omega_t, R_t) = -& log(P(R=1|\Omega_t))\cdot R_t \\
        	                   -& log(P(R=0|\Omega_t))\cdot (1 - R_t)
		\end{aligned}
		\end{equation}
where $\Omega_t$ stands for $o_1, o_2, .., o_t$ and $R_t = R$ is the final result of a match.
We simply set $R$ to be 1 if the player wins at the end and set it to be 0 otherwise.
\vspace{-1em}
\paragraph{Experiment Results}
The baseline network is trained on our dataset using global feature vector and evaluated with mean accuracy.
We report the results of all replays in Table~\ref{table:map}, serving as the baseline results for global state evaluation in MSC.
For Terran versus Terran matches, the mean accuracy in test set is 61.1\% after model converges. 
We also show the mean accuracies of different game phrases in Figure~\ref{fig:map}.
At the beginning of the game (0\%-25\%), it's hard to tell the probability of winning, as the mean accuracy of this curve is 52.9\%.
After half of the game (50\%-75\%), the mean accuracy could reach 64.2\%, while it's around 79.7\% at the end of the game (75\%-100\%).
\begin{table}
\begin{center}
\begin{tabular}{|l|c|c|c|c|c|c|}
\hline
V.S. & TvT & TvP & TvZ & PvP & PvZ & ZvZ \\
\hline
Baseline(\%) & 61.1 & 59.1 & 59.8 & 51.4 & 59.7 & 59.9 \\
\hline
\end{tabular}
\end{center}
\caption{\textbf{Mean accuracy for global state evaluation of all replays.}}
\label{table:map}
\end{table}
\begin{table}
\begin{center}
\begin{tabular}{|l|c|c|c|c|c|c|}
\hline
V.S. & TvT & TvP & TvZ & PvP & PvZ & ZvZ \\
\hline
Baseline(\%) & 74.1 & 74.8 & 73.5 & 76.3 & 75.1 & 76.1 \\
\hline
\end{tabular}
\end{center}
\caption{\textbf{Mean accuracy for build order prediction of all replays.}}
\label{table:acc}
\end{table}
\begin{table}
\begin{center}
\begin{tabular}{|l|c|c|c|c|c|c|}
\hline
Tasks & TvT & TvP & TvZ & PvP & PvZ & ZvZ \\
\hline
GSE & 50.9 & 57.0 & 56.1 & 57.8 & 56.9 & 54.7 \\
\hline
BOP & 73.1 & 69.6 & 74.8 & 74.2 & 74.2 & 74.9 \\
\hline
\end{tabular}
\end{center}
\caption{\textbf{Results for global state evaluation (GSE) and build order prediction (BOP) with both global and spatial features.}}
\label{table:accs}
\end{table}
\begin{figure}[t]
\begin{center}
	\includegraphics[width=\linewidth]{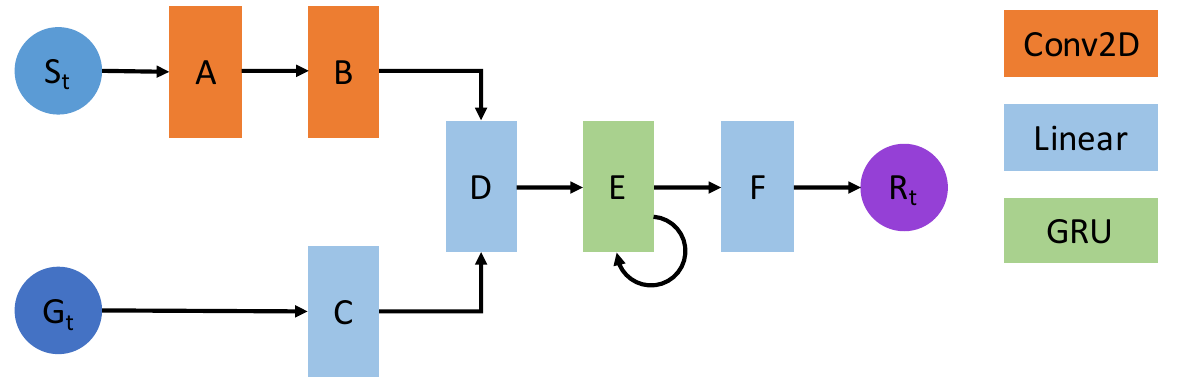}
\end{center}
	\caption{\textbf{Baseline network architecture for combining global and spatial features.} $S_t$ is the spatial feature tensor while $G_t$ is the global feature vector. A and B are convolution layers with channel size 16 and 32.    C, D and F are linear units with the number of units 128, 512 and $n_o$, while E is a GRU with size 128.}
	\label{fig:baseline_spatial}
\end{figure}
\subsection{Build Order Prediction}
\paragraph{Definition}
When playing StarCraft II, the most import task of macro-management is to decide what to train, build or research in the next step, given the current state from both sides. 
Such a task is named build order prediction in StarCraft community.
Formally, build order prediction is to predict the probability of a certain high-level action in the next step given current state at time step $t$, \ie predicting the value of $P(a_i|o_1, o_2, ..., o_t)$.
$a_i$ is one of the actions in the predefined high-level action space $A \cup \{a_\emptyset\}$.
\paragraph{Baseline Network}
The baseline network for build order prediction is similar to the one shown in Figure~\ref{fig:baseline}, except that the linear unit E produces $n_a$ outputs followed by a Softmax layer to produce $P(a_i|o_1, o_2, ..., o_t)$. $n_a$ is the number of actions in $A \cup \{a_\emptyset\}$.
\paragraph{Objective Function}
Negative Log Likelihood Loss (NLL) serves as our objective function, which is defined as Equation~\ref{equation:ce},
		\begin{equation}\label{equation:ce}
			J(\Omega_t, a_{t+1}) = - log(P(a_{t+1}|\Omega_t))
		\end{equation}
where $a_{t+1}$ is the action conducted in the next step in the replay file.
\paragraph{Experiment Results}
The baseline network is trained on our dataset using global feature vector and evaluated with Top-1 accuracy.
We report the results of all replays in Table~\ref{table:acc}, serving as the baseline results for build order prediction in MSC.
\subsection{Combine Global and Spatial Features}
We also design a baseline network for combining the global and spatial features, as shown in Figure~\ref{fig:baseline_spatial}.
The corresponding results for global state evaluation and build order prediction are reported in Table~\ref{table:accs} and serve as the baselines.
\subsection{Implementation Details}
Our algorithms are implemented using PyTorch\footnote{http://pytorch.org/}.
To train our baseline model, we use ADAM~\cite{kingma2014adam} for optimization and set learning rate to $0.001$.
At the end of every epoch, the learning rate is decreased by a factor of 2.
To avoid gradient vanishing or explosion, the batch size is set to 256, while the size of time steps is set to 20.
\section{Conclusion}
We released a new dataset MSC based on SC2LE, which focuses on macro-management in StarCraft II.
Different from the datasets in macro-management released before, we proposed a standard procedure for preprocessing, parsing and feature extraction.
We also defined the specifics of feature vector, the space of high-level actions and three subsets for training, validation and testing.
Our dataset preserves the high-level information directly parsed from replays as well as the final result (win or lose) of each match.
These characteristics make MSC the right place to experiment and evaluate various methods for assorted tasks in macro-management, such as build order prediction, global state evaluation and opening strategy clustering.
Among all these tasks, global state evaluation and build order prediction are two of the key tasks.
Thus, we proposed baseline models and presented initial baseline results for them.
However, other tasks require baselines as well, we leave these as future work and encourage other researchers to evaluate various tasks on MSC and report their results as baselines.

\small
\bibliographystyle{named}
\bibliography{ijcai18}

\end{document}